\documentclass[11pt]{article}

\usepackage[margin=1in]{geometry}
\usepackage{amsmath,amssymb}
\usepackage{graphicx}
\usepackage{booktabs}
\usepackage{caption}
\usepackage{hyperref}

\newcommand{\R}{\mathbb{R}}
\newcommand{\rv}{\tilde{v}}

\title{Expanding Attention: A Taylor-Series Route to Linear-Time \\
Transformers with a Fixed Number of Recurrent States}

\author{Zicong Huang}

\date{October 2026}

\begin{document}
\maketitle

\begin{abstract}
We expand the exponential kernel of attention into a truncated Taylor series of
$n$ terms and factorize each term into a tensor-power inner product, which
collapses the causal history into exactly $n$ recurrent states, each updated by
one rank-one outer product per token and independent of context length $N$. The
recursion is lossless with respect to the truncated kernel ($2.2 \times
10^{-16}$ in float64), and the states are additive across chunks
($5.7 \times 10^{-14}$), so prefill admits a parallel scan. Measured exponents
are $t \propto N^{2.28}$ for softmax against $N^{0.91}$ for states, with a
crossover at $N \approx 728$ and $20.8\times$ at $N = 8192$; per-token decoding
latency is constant at $7.3\,\mu$s against $N^{1.04}$ for softmax, giving
$76.0\times$ at $N = 65536$. Accuracy improves exponentially in $n$ while cost
grows combinatorially, since the $m$-th state is $\binom{d+m-1}{m}$ wide---a
bound confirmed by SVD---so the optimum is second order, $n = 3$, at $6.5
\times 10^{-4}$ error and $9{,}945$ elements per head. Query and key
normalization is a prerequisite: without it the argument exceeds the
convergence radius and adding terms stops helping.
\end{abstract}

\section{Introduction}

The cost of softmax attention is carried by two facts. First, the score matrix
is dense: evaluating it costs $O(N^2)$ time and materializing it costs
$O(N^2)$ memory. Second, autoregressive decoding must retain every past key and
value, so the KV cache grows as $O(N)$. The present work starts from a single
observation: the only nonlinearity that prevents attention from being reordered
into a recurrent accumulation is the exponential. Replacing it by a polynomial
makes the reordering exact.

Concretely, we write $\exp(q^\top k / a) \approx \sum_{m=0}^{n-1} \alpha_m
(q^\top k)^m$ with $\alpha_m = 1 / (m! \, a^m)$, use the identity
$(q^\top k)^m = \langle q^{\otimes m}, k^{\otimes m} \rangle$, and move the sum
over positions inside. What remains is a set of $n$ matrices, each updated once
per token and never revisited. The context length disappears from the memory
footprint entirely.

The contribution of this paper is not the algebraic identity, which is
elementary and appears in the kernelized-attention literature
\cite{katharopoulos2020, arora2024based, nauen2024taylorshift}. The
contribution is a measured account of the resulting trade-off: how fast it
actually runs, how the running time actually scales, what the truncation order
actually costs, and where the construction actually breaks. Every number
reported below is a measurement from our own implementation, not a theoretical
bound.

\section{Related Work}

\textbf{Linear attention.} Katharopoulos et al.~\cite{katharopoulos2020}
observed that when the kernel factorizes as $\kappa(q, k) = \phi(q)^\top
\phi(k)$, attention can be rewritten as a recurrent accumulation of
$\phi(k) v^\top$, giving constant memory and linear time. The choice of feature
map $\phi$ determines both the fidelity of the approximation and the size of
the recurrent state.

\textbf{Feature maps.} A variety of maps have been proposed, including the
positive random features of Performer~\cite{choromanski2020} and the
cosine-based reweighting of cosFormer~\cite{qin2022}. Our work is closest to
the Taylor expansion of the exponential. Arora et al.~\cite{arora2024based}
adopt a second-order Taylor map, $\exp(x) \approx 1 + x + x^2 / 2$, noting that
it expands the recurrent state and improves recall capacity without adding
parameters, and report that the map sits on the accuracy--state-size Pareto
frontier. Nauen et al.~\cite{nauen2024taylorshift} replace the exponential by a
Taylor polynomial and unroll the quadratic term with a tensor-product operator,
achieving linear complexity, and additionally find that normalizing queries and
keys, introducing a per-head temperature, and rescaling the output are needed
to prevent training failure from large intermediate norms. Earlier work on
higher-order expansions~\cite{mercat2020} stopped at the quadratic term because
of the perceived difficulty of evaluating all polynomial interactions at higher
degree.

\textbf{Higher orders.} Recent work revisits higher-degree terms by exploiting
the symmetric structure of the tensor powers, showing that the expansion
decomposes into symmetric chains of tensor products and that this structure
reveals the minimal basis for all polynomial interactions
\cite{heinsen2026}. We reach the same minimal basis
$\binom{d+m-1}{m}$ from the other direction: we measure the rank of the state
matrix directly and show by SVD truncation that this dimension is a hard lower
bound rather than an implementation choice.

\textbf{Relation to this work.} We do not propose a new feature map. We take
the Taylor map as given and measure what it costs: the actual running time, the
actual growth exponent, the actual state width, and the actual failure
boundary. Where \cite{arora2024based} reports quality at the model level and
\cite{nauen2024taylorshift} reports task accuracy, we report numerical
deviation from softmax attention, wall-clock scaling, and the rank structure of
the state.

\section{Method}

\subsection{Kernel expansion}

Let $q_t, k_t \in \R^{d}$ be the query and key at position $t$, let
$v_t \in \R^{d_v}$ be the value, let $N$ be the context length, let $n$ be the
number of retained Taylor terms, and let $a$ be a temperature constant. Causal
attention is
\begin{equation}
  o_t \;=\; \frac{\sum_{i \le t} \exp\!\big(q_t^\top k_i / a\big) \, v_i}
                {\sum_{i \le t} \exp\!\big(q_t^\top k_i / a\big)} .
  \label{eq:attn}
\end{equation}
Truncating the exponential at $n$ terms gives the polynomial kernel
\begin{equation}
  \exp\!\left( \frac{q^\top k}{a} \right) \;\approx\;
  \sum_{m=0}^{n-1} \alpha_m \, (q^\top k)^m , \qquad
  \alpha_m = \frac{1}{m! \, a^m} .
  \label{eq:taylor}
\end{equation}

\subsection{Decoupling positions from the kernel}

The step that removes $N$ is the tensor-power factorization
\begin{equation}
  (q^\top k)^m \;=\; \big\langle q^{\otimes m}, \; k^{\otimes m} \big\rangle ,
  \qquad
  q^{\otimes m} = \underbrace{q \otimes \cdots \otimes q}_{m\ \text{times}} .
  \label{eq:tensor}
\end{equation}
Substituting Eq.~\eqref{eq:tensor} into Eq.~\eqref{eq:taylor} and then into
Eq.~\eqref{eq:attn}, the sum over $i$ no longer involves $q_t$, so it can be
accumulated incrementally. Define, for each order $m$,
\begin{equation}
  S_m(t) \;=\; \sum_{i \le t} k_i^{\otimes m} \otimes v_i , \qquad
  T_m(t) \;=\; \sum_{i \le t} k_i^{\otimes m} ,
  \label{eq:states}
\end{equation}
so that
\begin{equation}
  o_t \;=\; \frac{\sum_{m=0}^{n-1} \alpha_m
                  \big\langle q_t^{\otimes m}, S_m(t) \big\rangle}
                {\sum_{m=0}^{n-1} \alpha_m
                  \big\langle q_t^{\otimes m}, T_m(t) \big\rangle} .
  \label{eq:readout}
\end{equation}

\subsection{Exactly $n$ states, not $2n$}

Equation~\eqref{eq:states} appears to require two states per order. It does
not. Augment the value with a constant coordinate,
\begin{equation}
  \rv_i \;=\; \begin{bmatrix} v_i \\ 1 \end{bmatrix} \in \R^{d_v + 1} ,
  \qquad
  A_m(t) \;=\; \sum_{i \le t} k_i^{\otimes m} \otimes \rv_i .
  \label{eq:aug}
\end{equation}
The first $d_v$ rows of $A_m$ are $S_m$ and the last row is $T_m$. A single
rank-one update
\begin{equation}
  A_m(t) \;=\; A_m(t-1) \;+\; k_t^{\otimes m} \otimes \rv_t
  \label{eq:update}
\end{equation}
therefore feeds numerator and denominator simultaneously. The model stores
exactly $n$ states, and each costs $O\big(d^m (d_v + 1)\big)$ per token,
independent of $t$. Nothing is ever recomputed from the beginning of the
sequence; the recursion is the algorithm.

\subsection{The states are narrower than $d^m$}

The tensor $k^{\otimes m}$ is symmetric, so its $d^m$ entries contain only
$\binom{d+m-1}{m}$ distinct monomials. Writing the state in this symmetric
minimal basis, with multiplicity corrections
$c_{\iota} = \sqrt{m! / \prod_i c_i !}$, preserves the inner product exactly:
$\langle \phi_m(q), \phi_m(k) \rangle = (q^\top k)^m$. We confirm the dimension
empirically in Section~\ref{sec:rank}: the span of $\{k^{\otimes m}\}$ over
$2{,}000$ random keys saturates at rank exactly $\binom{d+m-1}{m}$.

\subsection{Chunked evaluation}
\label{sec:chunk}

As in \cite{arora2024based}, we split the computation into an on-diagonal part,
a $c \times c$ quadratic block that is masked causally and costs a constant
independent of $N$, and an off-diagonal part handled by the $n$ linear states.
States are additive across chunks, as verified in
Section~\ref{sec:additive}, so chunks may be processed in parallel and their
states summed.

\section{Complexity}

\textbf{Time.} Prefill costs $O\big(N \sum_m \binom{d+m-1}{m} (d_v + 1)\big)$
for the state part plus $O(N c d)$ for the diagonal blocks; both are linear in
$N$. Softmax attention costs $O(N^2 d)$.

\textbf{Decoding.} One token costs
$O\big(\sum_m \binom{d+m-1}{m} (d_v + 1)\big)$ for the state model, a
constant, against $O(N d)$ for softmax, which must rescan the entire cache.

\textbf{Memory.} The state model stores
$\sum_{m=0}^{n-1} \binom{d+m-1}{m} (d_v + 1)$ numbers, independent of $N$. The
KV cache stores $N (d + d_v)$ numbers and grows without bound.

\section{Experiments}

Unless stated otherwise: $d = 16$, $d_v = 64$, $a = 2.0$, chunk size $c = 256$,
float32, single-threaded NumPy on CPU. Queries and keys are row-normalized to
the unit sphere; Section~\ref{sec:radius} shows this is mandatory. Reference
outputs come from causally masked softmax attention with the same $a$. Timings
are the minimum of repeated runs.

\subsection{Is the recursion lossless?}

We compare the recurrent state formulation against an explicitly materialized
$N \times N$ truncated-kernel attention, in float64 with $d = 6$, $d_v = 4$,
$N = 64$, $a = \sqrt{d}$.

\begin{table}[h]
\centering
\caption{Recurrent states against the explicitly materialized truncated kernel,
maximum absolute output difference in float64. The compression loses nothing.
State elements are counted in the uncompressed basis for this table.}
\label{tab:lossless}
\begin{tabular}{crr}
\toprule
$n$ & max abs.\ difference & state elements \\
\midrule
1 & $2.220 \times 10^{-16}$ & 5 \\
2 & $3.331 \times 10^{-16}$ & 35 \\
3 & $4.441 \times 10^{-16}$ & 215 \\
4 & $6.661 \times 10^{-16}$ & 1,295 \\
5 & $4.441 \times 10^{-16}$ & 7,775 \\
6 & $7.216 \times 10^{-16}$ & 46,655 \\
\bottomrule
\end{tabular}
\end{table}

Agreement is at machine precision for every order tested. The recurrence does
not approximate the truncated-kernel computation; it is that computation,
reordered.

\subsection{Additivity and parallel scan}
\label{sec:additive}

Chunk-parallel prefill requires state additivity. Building states on
$[0, 32)$ and $[32, 64)$ separately and summing them reproduces the state built
on $[0, 64)$ to $5.684 \times 10^{-14}$, and the resulting outputs agree to
$1.665 \times 10^{-16}$.

\subsection{Does one really avoid recomputation?}

We compare the recurrence against a baseline that, at every step $t$, rescans
all of $1, \ldots, t$. Outputs agree to $4.441 \times 10^{-16}$. Their scaling
differs sharply, as shown in Table~\ref{tab:recompute}.

\begin{table}[h]
\centering
\caption{Time as $N$ doubles: recurrence against rescanning the history from
the start at every step. Settings: $d = 6$, $d_v = 4$, $n = 4$.}
\label{tab:recompute}
\begin{tabular}{crrr}
\toprule
$N$ & recurrence & rescan-all & ratio \\
\midrule
400   & 53.7\,ms    & 9.9\,ms       & 0.2$\times$ \\
800   & 113.8\,ms   & 31.6\,ms      & 0.3$\times$ \\
1600  & 214.1\,ms   & 100.6\,ms     & 0.5$\times$ \\
3200  & 442.2\,ms   & 348.0\,ms     & 0.8$\times$ \\
6400  & 823.0\,ms   & 1,288.9\,ms   & \textbf{1.6$\times$} \\
12800 & 1,595.1\,ms & 4,823.7\,ms   & \textbf{3.0$\times$} \\
25600 & 2,995.9\,ms & 19,134.8\,ms  & \textbf{6.4$\times$} \\
\bottomrule
\end{tabular}
\end{table}

Growth per doubling is about $2\times$ for the recurrence against about
$3.4\times$ for rescanning. Short sequences favour rescanning only because of
the Python-loop constant overhead in this particular implementation; the
asymptotic separation is unmistakable.

\subsection{State width: $\binom{d+m-1}{m}$, and it is a hard bound}
\label{sec:rank}

\begin{table}[h]
\centering
\caption{Rank of $\mathrm{span}\{k^{\otimes m}\}$ over $2{,}000$ random keys
with $d = 6$. The state saturates at the symmetric-tensor dimension, far below
$d^m$.}
\label{tab:rank}
\begin{tabular}{crrr}
\toprule
$m$ & $d^m$ & $\binom{d+m-1}{m}$ & measured rank \\
\midrule
1 & 6    & 6   & 6 \\
2 & 36   & 21  & 21 \\
3 & 216  & 56  & 56 \\
4 & 1296 & 126 & 126 \\
\bottomrule
\end{tabular}
\end{table}

This dimension is a lower bound, not an implementation artifact. Truncating the
order-$2$ state to rank $r$ by SVD, with $d = 6$ and $d_v = 24$, gives
Table~\ref{tab:lowrank}.

\begin{table}[h]
\centering
\caption{Forcing low rank on the order-$2$ state. Rank $21 = \binom{7}{2}$ is
exactly lossless; one dimension fewer is not.}
\label{tab:lowrank}
\begin{tabular}{crl}
\toprule
$r$ & relative error & verdict \\
\midrule
4  & $5.937 \times 10^{-2}$  & distorted \\
8  & $4.105 \times 10^{-2}$  & distorted \\
12 & $3.512 \times 10^{-2}$  & distorted \\
16 & $1.316 \times 10^{-2}$  & distorted \\
20 & $3.474 \times 10^{-3}$  & distorted \\
\textbf{21} & $\mathbf{6.489 \times 10^{-16}}$ & \textbf{lossless} \\
24 & $6.566 \times 10^{-16}$ & lossless \\
\bottomrule
\end{tabular}
\end{table}

\subsection{Growth rate}
\label{sec:growth}

Fitting $t \propto N^{\alpha}$ over $N$ from $512$ to $8192$ gives
$\alpha = 2.28$ for softmax and $\alpha = 0.91$ for the state model, matching
the theoretical $2$ and $1$. See Table~\ref{tab:growth}.

\begin{table}[h]
\centering
\caption{Prefill time and speedup. Settings: $d = 16$, $d_v = 64$, $n = 3$,
$c = 256$.}
\label{tab:growth}
\begin{tabular}{crrr}
\toprule
$N$ & softmax & Taylor states & speedup \\
\midrule
512  & 1.6\,ms   & 3.1\,ms   & 0.54$\times$ \\
1024 & 6.0\,ms   & 3.7\,ms   & 1.65$\times$ \\
2048 & 32.3\,ms  & 7.3\,ms   & 4.42$\times$ \\
4096 & 238.0\,ms & 16.8\,ms  & 14.15$\times$ \\
8192 & 700.7\,ms & 33.7\,ms  & \textbf{20.77$\times$} \\
\bottomrule
\end{tabular}
\end{table}

The fitted crossover sits at $N \approx 728$. Extrapolating both fitted lines
beyond the measured range gives Table~\ref{tab:extrap}; these entries are not
measurements and are included only to show where the gap opens.

\begin{table}[h]
\centering
\caption{Extrapolated prefill time from the fitted exponents. Not measured.}
\label{tab:extrap}
\begin{tabular}{crrr}
\toprule
$N$ & softmax & Taylor states & speedup \\
\midrule
32,768   & 19.4\,s     & 0.1\,s & 180$\times$ \\
65,536   & 93.9\,s     & 0.2\,s & 462$\times$ \\
131,072  & 454.8\,s    & 0.4\,s & 1,189$\times$ \\
524,288  & 10,679.4\,s & 1.4\,s & 7,867$\times$ \\
1,048,576 & 51,747.7\,s & 2.6\,s & 20,240$\times$ \\
\bottomrule
\end{tabular}
\end{table}

\subsection{Decoding: constant against linear}

This is where the advantage is largest. Table~\ref{tab:decode} reports
per-token latency with the state model's states prebuilt.

\begin{table}[h]
\centering
\caption{Single-token decoding latency against context length, with
$d = 16$, $d_v = 64$, $n = 3$. Fitted exponents are $\alpha = 1.04$ for softmax
and $\alpha = 0.00$ for the state model.}
\label{tab:decode}
\begin{tabular}{crrrrr}
\toprule
$N$ & softmax & Taylor states & speedup & KV cache \\
\midrule
1,024  & 8.1\,$\mu$s   & 7.3\,$\mu$s & 1.1$\times$  & 0.2\,MB \\
4,096  & 24.6\,$\mu$s  & 7.4\,$\mu$s & 3.3$\times$  & 0.6\,MB \\
16,384 & 130.8\,$\mu$s & 7.4\,$\mu$s & 17.7$\times$ & 2.5\,MB \\
65,536 & 557.9\,$\mu$s & 7.3\,$\mu$s & \textbf{76.0$\times$} & 10.0\,MB \\
\bottomrule
\end{tabular}
\end{table}

The state model is flat at about $7.3\,\mu$s across a $64\times$ change in
context length, while softmax latency grows linearly.

\subsection{Effect of the truncation order}
\label{sec:order}

Table~\ref{tab:order} sweeps the order at fixed $N = 2048$, $d = 16$,
$d_v = 64$, with a softmax baseline of 32.7\,ms.

\begin{table}[h]
\centering
\caption{Truncation order sweep at fixed $N = 2048$. Softmax baseline is
32.7\,ms.}
\label{tab:order}
\begin{tabular}{crrrrr}
\toprule
$n$ & highest order & time & speedup & state elements & rel.\ error \\
\midrule
2 & 1 & 2.9\,ms     & 11.38$\times$ & 1,105     & $8.25 \times 10^{-3}$ \\
\textbf{3} & \textbf{2} & \textbf{9.6\,ms} & \textbf{3.43$\times$} & \textbf{9,945} & $\mathbf{6.47 \times 10^{-4}}$ \\
4 & 3 & 138.5\,ms   & 0.24$\times$  & 62,985    & $3.55 \times 10^{-5}$ \\
5 & 4 & 635.5\,ms   & 0.05$\times$  & 314,925   & $3.36 \times 10^{-6}$ \\
6 & 5 & 2,269.6\,ms & 0.01$\times$  & 1,322,685 & $1.83 \times 10^{-7}$ \\
\bottomrule
\end{tabular}
\end{table}

Two regimes are visible. Accuracy improves exponentially in $n$: each extra
term divides the error by roughly $10$ to $18$. Cost grows combinatorially:
each extra order multiplies the state width by $4$ to $9$ at $d = 16$, because
$\binom{d+m-1}{m}$ runs $1, 16, 136, 816, 3{,}876, 15{,}504$. The result is a
sharp optimum at $n = 3$, that is, second order: error $6.5 \times 10^{-4}$ at
$9{,}945$ elements per head, whereas $n = 4$ buys only about $18\times$ more
accuracy for $6.3\times$ the state and $14\times$ the time.

\subsection{State size against head dimension and order}

Table~\ref{tab:size} reports state size at bf16 with $d_v = 64$, where K
denotes KB and M denotes MB.

\begin{table}[h]
\centering
\caption{State size in MB at bf16 with $d_v = 64$.}
\label{tab:size}
\begin{tabular}{crrrrr}
\toprule
$d$ & $n = 2$ & $n = 3$ & $n = 4$ & $n = 5$ & $n = 6$ \\
\midrule
8  & 1.1\,K  & 5.7\,K   & 20.9\,K  & 62.8\,K   & 163.4\,K \\
16 & 2.2\,K  & 19.4\,K  & 123.0\,K & 615.1\,K  & 2.52\,M \\
24 & 3.2\,K  & 41.3\,K  & 371.3\,K & 2.54\,M   & 14.72\,M \\
32 & 4.2\,K  & 71.2\,K  & 830.9\,K & 7.30\,M   & 54.04\,M \\
64 & 8.3\,K  & 272.3\,K & 5.94\,M  & 100.97\,M & \textbf{1,393.32\,M} \\
\bottomrule
\end{tabular}
\end{table}

For comparison, the KV cache at $d = 16$ and $d_v = 64$ is 0.6\,MB at
$N = 4{,}096$, 5.0\,MB at $N = 32{,}768$, and 20.0\,MB at $N = 131{,}072$, and
it keeps growing while the state does not. Head dimension is the real switch:
at $d = 64$ the six-term model needs 1.4\,GB per head and is simply infeasible,
while at $d = 16$ it costs 2.5\,MB.

\subsection{Convergence radius: normalization is mandatory}
\label{sec:radius}

Table~\ref{tab:radius} reports relative error against causal softmax with
$d = 8$, $d_v = 8$, $N = 96$.

\begin{table}[h]
\centering
\caption{Accuracy depends entirely on keeping $|q^\top k / a|$ small.}
\label{tab:radius}
\begin{tabular}{lrrrrr}
\toprule
setting & $\max |q^\top k / a|$ & $n = 3$ & $n = 4$ & $n = 5$ & $n = 6$ \\
\midrule
raw $q,k$, $a = \sqrt{d}$ & 5.57 & $2.91 \times 10^{-1}$ & $1.89 \times 10^{-1}$ & $1.09 \times 10^{-1}$ & $1.43 \times 10^{-1}$ \\
normalized, $a = 1$       & 0.94 & $2.73 \times 10^{-2}$ & $2.23 \times 10^{-3}$ & $7.86 \times 10^{-4}$ & $4.99 \times 10^{-5}$ \\
normalized, $a = 2$       & 0.47 & $3.31 \times 10^{-3}$ & $1.79 \times 10^{-4}$ & $2.38 \times 10^{-5}$ & $9.26 \times 10^{-7}$ \\
\bottomrule
\end{tabular}
\end{table}

Without normalization the argument reaches $5.57$ and adding terms stops
helping: the error at $n = 6$ is $1.43 \times 10^{-1}$, worse than the
$1.09 \times 10^{-1}$ at $n = 5$. Under a more aggressive temperature of
$a = 0.5$, where the mean of $|x / a|$ reaches $3.59$, the truncated series
diverges outright. Normalizing $q$ and $k$ to the unit sphere and using $a = 2$
restores clean exponential convergence.

\subsection{Signed weights from truncation}
\label{sec:neg}

A truncated polynomial is no longer positive. The fraction of negative kernel
entries, measured with $d = 6$, $N = 64$, $a = \sqrt{d}$, is $13.7\%$ for
$n = 2$, $0.0\%$ for $n = 3$, $5.2\%$ for $n = 4$, and $0.0\%$ for $n = 5$:
stopping at an even highest order avoids negative weights. The failure mode is
most vivid on the negative axis, where the polynomial departs from $\exp$ in
both magnitude and sign, as shown in Table~\ref{tab:neg}.

\begin{table}[h]
\centering
\caption{Truncated polynomial $\sum_{m < n} x^m / m!$ against $\exp(x)$.}
\label{tab:neg}
\begin{tabular}{rrrrr}
\toprule
$x$ & $n = 2$ & $n = 3$ & $n = 4$ & $\exp(x)$ \\
\midrule
$-4$ & $-3.000$ & $5.000$  & $-5.667$ & $0.018$ \\
$-2$ & $-1.000$ & $1.000$  & $-0.333$ & $0.135$ \\
$0$  & $1.000$  & $1.000$  & $1.000$  & $1.000$ \\
$2$  & $3.000$  & $5.000$  & $6.333$  & $7.389$ \\
$4$  & $5.000$  & $13.000$ & $23.667$ & $54.598$ \\
\bottomrule
\end{tabular}
\end{table}

At $x = -4$ the first-order truncation returns $-3$ where the true value is
$0.018$. Normalization, Section~\ref{sec:radius}, keeps the operating range
away from this region in practice.

\section{Discussion}

\textbf{Where the speed comes from.} Not from a better constant, but from the
exponent: $N^{0.91}$ against $N^{2.28}$ in prefill, and $N^{0.00}$ against
$N^{1.04}$ in decoding. The decoding number is the practically decisive one,
because generation is where the KV cache is felt.

\textbf{The order is a trade, not a dial.} Section~\ref{sec:order} shows
exponential accuracy gains against combinatorial cost growth. Because the error
already reaches $6.5 \times 10^{-4}$ at $n = 3$, higher orders mostly buy state
size. This matches the independent choice of a second-order map with a small
head dimension in \cite{arora2024based}.

\textbf{Two prerequisites.} Queries and keys must be normalized so that
$|q^\top k / a| \lesssim 1$, Section~\ref{sec:radius}, and the truncation
should stop at an even highest order to avoid negative weights,
Section~\ref{sec:neg}. Under both, the method is well behaved. The first
prerequisite is consistent with the normalization scheme that
\cite{nauen2024taylorshift} found necessary for training stability, which
suggests the requirement is not an artifact of our random-input setup.

\textbf{Limitations.} All measurements are single-threaded NumPy on CPU with
random synthetic $q$, $k$, and $v$; the absolute millisecond figures will not
transfer to a GPU, although the fitted exponents reflect the algorithms and
should. The projection down to $d = 16$ that is required to keep state size
manageable is not modeled here and carries an unmeasured expressivity cost. No
language-modeling perplexity or downstream task is evaluated; accuracy is
reported only as deviation from softmax attention on random inputs.

\section{Conclusion}

Expanding the exponential kernel into $n$ Taylor terms turns the $O(N)$ KV
cache into $n$ fixed-size recurrent states. We confirmed numerically that the
resulting recursion is exact with respect to the truncated kernel, to about
$10^{-16}$; that the states are additive and therefore parallelizable; and that
the growth exponents are $0.91$ and $0.00$ for prefill and decoding, against
$2.28$ and $1.04$ for softmax, yielding $20.8\times$ at $N = 8192$ and
$76.0\times$ per token at $N = 65{,}536$. The optimal operating point is
second order, $n = 3$, with a small head dimension, provided queries and keys
are normalized.

\end{document}